\documentclass[10pt,twocolumn,letterpaper]{article}

\usepackage{iccv}      
\usepackage{multirow}
\usepackage{makecell}
\usepackage{tabularx}
\usepackage{amsmath}
\usepackage{amssymb}
\usepackage{pifont}
\usepackage{amsmath}
\usepackage{graphicx}
\usepackage{algorithm}
\usepackage{algorithmic}
\usepackage{subcaption}
\usepackage[accsupp]{axessibility}
\definecolor{iccvblue}{rgb}{0.21,0.49,0.74}
\usepackage[pagebackref,breaklinks,colorlinks,allcolors=iccvblue]{hyperref}

\def\paperID{7864} 
\def\confName{ICCV}
\def\confYear{2025}

\title{One Polyp Identifies All: One-Shot Polyp Segmentation with SAM via Cascaded Priors and Iterative Prompt Evolution}

\author{Xinyu Mao$^{1}$ \hspace{2mm} Xiaohan Xing$^{\dagger,2}$ \hspace{2mm} Fei Meng$^{3}$ \hspace{2mm}Jianbang Liu$^{1}$ \hspace{2mm} Fan Bai$^{1}$ \hspace{2mm} Qiang Nie$^{4}$ \hspace{2mm} Max Meng$^{\dagger,5}$\\
 $^{1}$Chinese University of Hong Kong\hspace{4mm} 
  $^{2}$Stanford University\hspace{4mm} \\
  $^{3}$Hong Kong University of Science and Technology\hspace{4mm} \\
  $^{4}$Hong Kong University of Science and Technology(Guangzhou)\hspace{4mm} \\
  $^{5}$Southern University of Science and Technology\\
  {\tt\small xhxing@stanford.edu, max.meng@ieee.org}
}

\begin{document}
\maketitle
\renewcommand\thefootnote{}
\footnotetext{$^\dagger$Corresponding author.}
\footnotetext{$^*$Project page: \url{https://github.com/Hectormxy/OP-SAM}}
\setcounter{footnote}{0}

\begin{abstract}
Polyp segmentation is vital for early colorectal cancer detection, yet traditional fully supervised methods struggle with morphological variability and domain shifts, requiring frequent retraining. Additionally, reliance on large-scale annotations is a major bottleneck due to the time-consuming and error-prone nature of polyp boundary labeling. Recently, vision foundation models like Segment Anything Model (SAM) have demonstrated strong generalizability and fine-grained boundary detection with sparse prompts, effectively addressing key polyp segmentation challenges. However, SAM’s prompt-dependent nature limits automation in medical applications, since manually inputting prompts for each image is labor-intensive and time-consuming. We propose OP-SAM, a \textbf{O}ne-shot \textbf{P}olyp segmentation framework based on \textbf{SAM} that automatically generates prompts from a single annotated image, ensuring accurate and generalizable segmentation without additional annotation burdens. Our method introduces Correlation-based Prior Generation (CPG) for semantic label transfer and Scale-cascaded Prior Fusion (SPF) to adapt to polyp size variations as well as filter out noisy transfers. Instead of dumping all prompts at once, we devise Euclidean Prompt Evolution (EPE) for iterative prompt refinement, progressively enhancing segmentation quality. Extensive evaluations across five datasets validate OP-SAM’s effectiveness. Notably, on Kvasir, it achieves 76.93\% IoU, surpassing the state-of-the-art by 11.44\%$^*$.
\end{abstract}    
\section{Introduction}
\label{sec:introduction}



\begin{figure}[h]
\centering
\includegraphics[width=0.45\textwidth]{{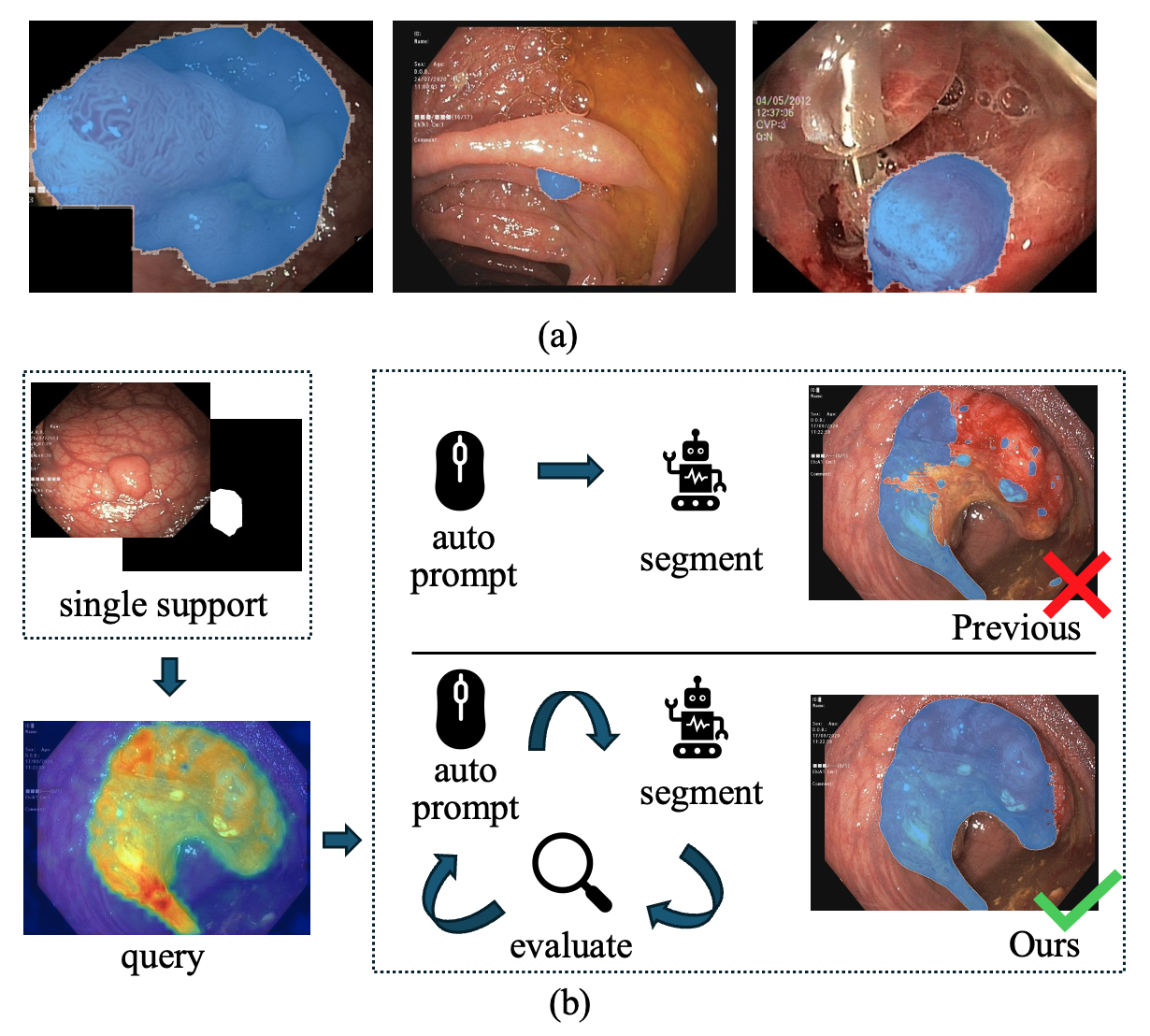}}
\label{fig:meth_comp}
\caption{Challenges in SAM-based polyp segmentation and methods comparison. (a) The size variations of polyps and interference like bubbles and reflections complicate polyp perception. (b) Previous methods pour all the prompts simultaneously, while our method introduces evaluation to add prompts interactively.}
\end{figure}

Colorectal cancer (CRC), ranking as the third most prevalent cancer globally \cite{jain2023coinnet}, begins when small, benign cell clusters called polyps form on the colon's interior lining. Early polyp detection via colonoscopy can prevent CRC, yet polyp segmentation, especially with fully supervised methods, faces two key limitations. First, polyp segmentation is inherently challenging due to significant variations in size, color, and texture, requiring extensive expert annotations to enhance the segmentation capability \cite{rahman2023medical}. However, ambiguous boundaries make precise and large-scale annotations time-consuming and prone to errors, which can inversely degrade model performance. Second, clinical deployment of polyp segmentation models requires strong generalizability across datasets. However, domain shifts from variations in endoscopic devices and patient demographics hinder transferability. Fully supervised models tend to overfit the training data, resulting in poor generalization and requiring frequent, resource-intensive retraining, which is clinically inefficient and impractical \cite{tian2023self}. Therefore, a polyp segmentation model is needed that minimizes annotation dependence while ensuring accurate and generalizable segmentation.

Vision foundation models (VFM), such as CLIP \cite{radford2021clip}, DINOv2 \cite{oquab2023dinov2}, and SAM2 \cite{ravi2024sam}, are trained on large-scale datasets, endowing them with strong transferability. Among them, SAM2 excels in segmentation with its fine-grained contour capture and impressive zero-shot capabilities, requiring only sparse prompts like points or bounding boxes. This makes it ideal for polyp segmentation with precise boundary detection, better generalization, and reduced annotation dependency. However, SAM’s reliance on manual prompts limits its clinical feasibility. Manually providing prompts for each image is time-consuming and lacks automation. This prompts a key question: Can a single annotated image automatically generate reliable prompts for SAM, ensuring accuracy and generalization without additional annotation burdens?

Recent research has explored integrating SAM with single-shot semantic segmentation to enable automated segmentation using only one annotation. The pioneer work PerSAM \cite{zhang2023personalize} uses prototype-based matching for prompt selection. Matcher \cite{liu2023matcher} adopts a prompt-free pixel-level feature matching approach, and ProtoSAM \cite{ayzenberg2024protosam} improves localization with multi-scale features. However, these methods have notable limitations in polyp segmentation. First, these methods struggle to handle morphological variations, particularly differences in polyp size (see Fig. 1(a)). Since a single support image provides limited information, it may fail to capture the entire polyp region when the query is too large, while being prone to false positives for smaller query polyps. Second, previous approaches dump all prompts into SAM simultaneously (see Fig. 1(b)), leading to a trade-off: too few prompts fail to provide sufficient information, whereas excessive prompts introduce noise, degrading performance. Thus, an iterative strategy for optimal prompt selection and placement is required.

In this work, we introduce OP-SAM, a flexible, feedback-driven automatic prompting segmentation framework with just \textbf{O}ne \textbf{P}olyp mask and \textbf{SAM}. Our approach handles lesion size variations by adaptively extracting multi-scale semantics from a single support image and iteratively placing prompts for optimal segmentation. First, to achieve accurate label transfer from the support image and capture complete polyp information, we propose \textit{Correlation-based Prior Generation} (CPG), which enhances feature-matching accuracy by leveraging patch-wise feature correlation cross- and within-images. Second, to handle polyp size variability, we introduce multi-scale prior and reverse-transferring adaptive fusion, namely \textit{Scale-cascaded Prior Fusion} (SPF). Multi-granularity priors are adaptively fused based on reverse transfer to eliminate false-positive noise. Additionally, we develop \textit{Euclidean Prompt Evolution} (EPE), an algorithm inspired by SAM’s interactive training pipeline, utilizing Euclidean distance transform to refine segmentation based on output from the previous round iteratively. Given a precise prior, we establish a cyclic process of prompting, segmentation, and evaluation, ensuring comprehensive polyp coverage.

Extensive evaluation across multiple regional five datasets—Kvasir \cite{jha2020kvasir}, PolypGen \cite{ali2023multi}, CVC-ClinicDB \cite{bernal2015wm}, CVC-ColonDB \cite{tajbakhsh2015automated} and Piccolo \cite{sanchez2020piccolo}—demonstrates significant improvements over current state-of-the-art (SOTA) methods, with maximal IoU increases of 11.44\%. Notably, on the Kvasir dataset, our approach even surpasses oracle performance which uses ground truth from test images to generate prompts. We further test OP-SAM on a manually curated dataset containing extreme-sized polyps. Our approach shows a 10.26\% IoU improvement over existing methods, demonstrating robust performance in challenging yet clinically critical cases. Our key contributions can be summarized as follows:
\begin{itemize}
    \item We propose OP-SAM, an innovative training-free framework for one-shot polyp segmentation, leveraging VFM semantics to automate prompt generation for SAM, enabling efficient adaptation with minimal annotations.
    
    \item We propose the CPG and SPF modules to generate fine-grained semantic priors, effectively handling polyp size variability. Additionally, we introduce the EPE algorithm, which simulates human-like interactive prompting to enhance segmentation accuracy.

    \item Through comprehensive experiments across diverse datasets, we demonstrate the remarkable performance and generalization ability of our methods over state-of-the-art methods. In certain datasets, it even surpasses the oracle method where ground truths are given.
\end{itemize}

\section{Related Work}
\begin{figure*}[t]
  \centering
   \includegraphics[width=0.98\textwidth]{{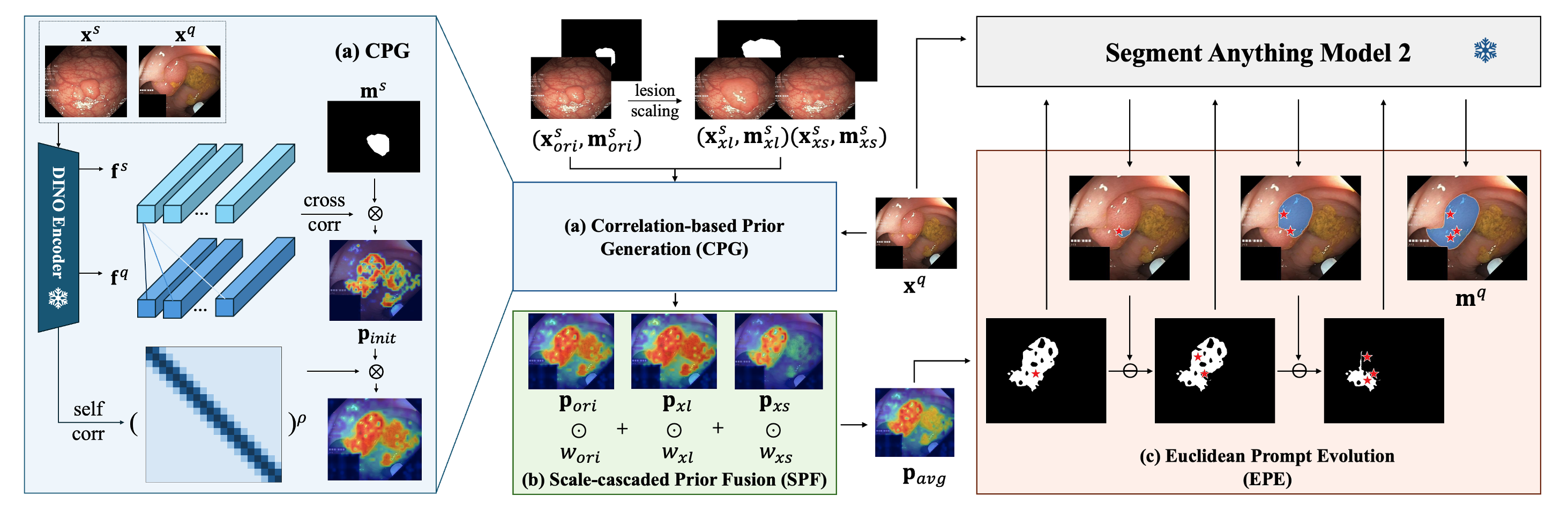}}
    \caption{The overall framework of OP-SAM. After lesion scaling, the support image $\mathbf{x}^{s}_{ori}$ with its augmentation (extra-large lesion $\mathbf{x}^{s}_{xl}$, extra-small lesion $\mathbf{x}^{s}_{xs}$) is fed forward into CPG. SPF then adaptively fuses the generated multiple prior $\mathbf{p}_{ori}, \mathbf{p}_{xl}, \mathbf{p}_{xs}$. Finally, EPE interactively adds prompts after evaluating the last-round prediction. The details of CPG are illustrated on the left. Coarse prior is generated with cross-correlation, followed by iterative self-correlation refinement.}
   \label{fig:frame}
\end{figure*}
\label{sec:formatting}
\noindent{\textbf{Polyp Segmentation}}
The accurate segmentation of polyps remains a significant challenge in medical image analysis, primarily due to the complex intestinal environment and diverse polyp appearances. Traditional approaches have addressed these challenges through various computational strategies, including multi-scale feature \cite{msrf}, shallow attention \cite{wei2021shallow}, cascaded attention \cite{rahman2023medical}, boundary information \cite{zhou2023cross} and so on \cite{wang2023xbound, patel2022fuzzynet, jain2023coinnet,jha2024transnetr}. Recognizing the resource-intensive nature of expert annotation, researchers have explored alternative learning paradigms such as semi-supervised \cite{li2022tccnet}, weakly-supervised \cite{wang2023s} or self-supervised \cite{tian2023self} approaches. While recent adaptations of the Segment Anything Model \cite{ravi2024sam} for polyp segmentation \cite{li2024polyp, rahman2024pp} show promise, they still require model fine-tuning and manual intervention. Our approach overcomes these limitations by providing a training-free, fully automated system with minimal supervision—requiring only a single image-mask pair.

\noindent{\textbf{Few Shot Segmentation}}
Few-shot semantic segmentation (FSSS) addresses the challenge of object segmentation with limited labelled examples. Current approaches primarily fall into two categories: prototype-based and matching-based methods. The former \cite{yang2023mianet, lang2022learning, li2021adaptive, chen2021semantically} generates representative feature descriptors through average pooling while the latter \cite{shi2022dense, liu2022intermediate, zhang2021few, gao2022mutually} leverages pixel-wise correlations between support and query features. Despite their effectiveness within specific domains, these methods often struggle with domain shifts. In view of this problem, some methods \cite{nie2024cross, lei2022cross} propose cross-domain few-shot segmentation to tackle out-of-domain performance degradation. SegGPT \cite{wang2023seggpt} attempts to address this through in-context learning. However, it trades accuracy for robustness.

\noindent{\textbf{SAM-based Segmentation}}
Recently, SAM \cite{ravi2024sam}, an interactive segmentation model, has demonstrated impressive zero-shot abilities with prompt inputs such as points, boxes and coarse masks. However, its semantic-agnostic nature and reliance on manual prompting limit practical applications. Following methods like Semantic-SAM \cite{li2023semantic} and MedSAM \cite{ma2024segment} have focused on incorporating semantic knowledge through fine-tuning. Recently, several methods have introduced SAM into FSSS, using either prototype comparison (PerSAM \cite{zhang2023personalize}, ProtoSAM \cite{ayzenberg2024protosam}) or pixel-matching (Matcher \cite{liu2023matcher}). These approaches, however, overlook appearance variations between query and support images and employ rigid prompting strategies. Our method addresses these limitations through refined prior generation and adaptive, feedback-driven prompting.
\section{Method}

OP-SAM is a training-free one-shot polyp segmentation framework with a semantically unaware segmentation model (SAM2 \cite{ravi2024sam}). Formally, given a support image $\mathbf{x}^{s}$ with its ground-truth mask ${\mathbf{m}}^{s}$, the goal is to segment a binary mask ${\mathbf{\Tilde{m}}}^{q}$ for any query image $\mathbf{x}^{q}$ containing polyps. The overall framework is illustrated in \cref{fig:frame}. For query image $\mathbf{x}^{q}$, the CPG module examines feature relationships both across and within images to establish coarse semantic context. Then, SPF adaptively fuse augmented versions of the query image to ensure robust detection across size variations. With accurate prior output from SPF, EPE implements an iterative refinement process, generating and adjusting segmentation prompts based on continuous feedback between the SAM2 output and semantic prior.

\subsection{Correlation-based Prior Generation}

We propose CPG (see Fig. 2(a)) to transfer the semantic label from support to query based on cross-image correlation. We employ the frozen image encoder DINOv2 \cite{oquab2023dinov2} to extract semantic features for both support and query images. Given input $\mathbf{x}^{s}$ and $\mathbf{x}^{q}\in \mathbb{R}^{\mathrm{H}\times\mathrm{W}\times\mathrm{3}}$, the encoder produces patch-level feature embeddings $\mathbf{f}^{s}, \mathbf{f}^{q} \in \mathbb{R}^{\mathrm{hw}\times\mathrm{D}}$. Then the cross-correlation matrix $\mathbf{S}_{corr}\in\mathbb{R}^{\mathrm{hw}\times\mathrm{hw}}$ between query and support is calculated as,
\begin{equation}
    \centering    \mathbf{S}_{corr}=\frac{\mathbf{f}^{q}\cdot{\mathbf{f}^{s}}^{T}}{\sqrt{D}}.
\end{equation}
Matrix $\mathbf{S}_{corr}$ quantifies patch-level similarities between images. We resize the support mask $\mathbf{m}^{s}\in\mathbb{R}^{\mathrm{H}\times\mathrm{W}}$ to the feature level and flatten it into $\mathbf{m}^{s}_{r}\in\mathbb{R}^{\mathrm{h}\mathrm{w}\times1}$. The multiplication of $\mathbf{S_{corr}}$ and $\mathbf{m}^{s}_{r}$ gives the initial prior of the query by transferring the label from support to query through semantic closeness between respective patches.

However, the initial prior is too rough and discret as in \cref{fig:frame} $\mathbf{p}_{init}$ since polyp features inevitably vary across different images. To find a more comprehensive polyp prior, we draw inspiration from \cite{lin2023clip}, extract the self-correlation matrix from the last attention block in DINO encoder. Then we apply Sinkhorn normalization \cite{lin2023clip} (alternatively perform row and column normalization) to turn the self-correlation matrix into a symmetric one, facilitating following multi-round refinement. We vectorize the prior, compute the product of query-to-query patch-wise feature similarity $\mathbf{S}_{self}\in\mathbb{R}^{\mathrm{hw}\times\mathrm{hw}}$ and the coarse prior $\mathrm{vec}(\mathbf{p_{init}})=\mathbf{S}_{corr}\cdot\mathbf{m}^{s}_{r}\in\mathbb{R}^{\mathrm{hw}}$. For a given patch, its semantic response is refined by weighing its similarity with all other patches against the coarse prior distribution. This operation effectively refines and enhances the semantic prior. Self-refinement is performed in several $\rho$ runs. To sum up, the fine-grained prior-generation process can be formulated as:
\begin{equation}
    \mathbf{p}_{ori}={[\mathbf{S}_{self}]}^{\rho}\cdot\mathbf{S}_{corr}\cdot\mathbf{m}^{s}_{r}.
\end{equation}

\subsection{Scale-Cascaded Prior Fusion}
A single support image with fixed-size polyp struggles to provide robust guidance for query images with diverse polyp sizes. False-positive and false-negative misclassifications are common when a significant gap exists between query and support in lesion size. Considering the variation in polyp size, we zoom in and zoom out the lesion area of the support, respectively. The corresponding segmentation masks are scaled proportionally. Together with the original image-mask pairs $(\mathbf{x}^{s}_{ori}, \mathbf{m}^{s}_{ori})$, we feed forward the augmented version $(\mathbf{x}^{s}_{xl}, \mathbf{m}^{s}_{xl})$ and $(\mathbf{x}^{s}_{xs}, \mathbf{m}^{s}_{xs})$ into the CPG module which gives in total three priors $\mathbf{p}_{ori}, \mathbf{p}_{xl}$ and $\mathbf{p}_{xs}$. The enlarged prior $\mathbf{p}_{xl}$ typically captures broader lesion areas, while the shrinked prior $\mathbf{p}_{xs}$ emphasizes prominent lesion features.

\begin{figure}[t]
  \centering
   \includegraphics[width=0.45\textwidth]{{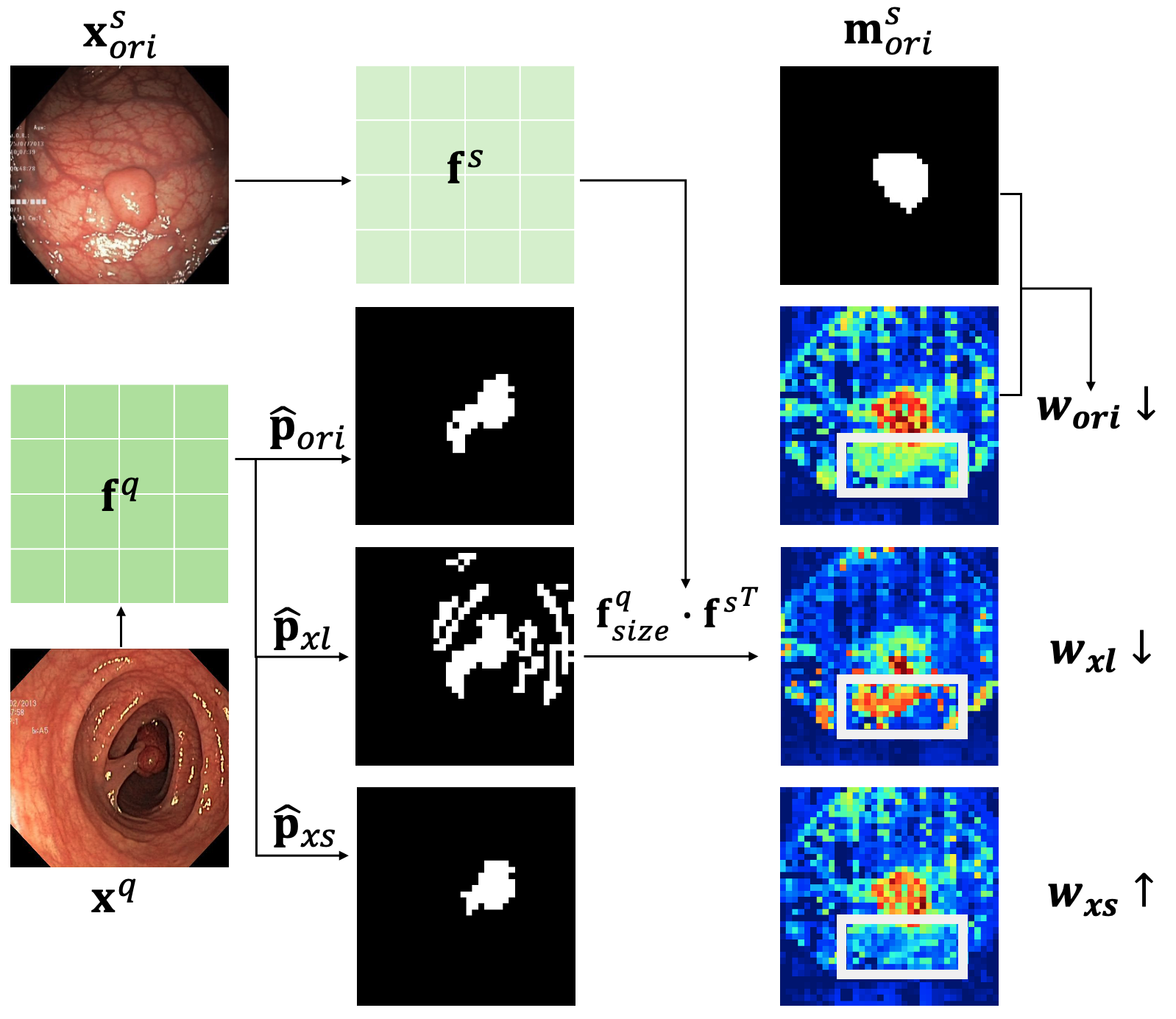}}
    \caption{The pipeline of SPF. It reversely performs prior generation from query to support, using features selected by binary query prior($\mathbf{f}^{q}_{size}$ denotes features selected by $\hat{\mathbf{p}}_{size}$ for $size \in(ori, xl, xs)$). Noisy query prior yields artefacts (highlighted by bounding box) outside the polyp region in support image.}
   \label{fig:sel}
\end{figure}

The scaling operation introduces a new problem: with three priors available, we need to select the most accurate one without ground truth. Hence, we propose a reverse transfer mechanism to assess prior quality, leveraging the known segmentation of the support image. Specifically, for each prior $\mathbf{p}_{size}, size\in(ori, xs, xl)$, we threshold it by $\tau$ and get a binary mask $\hat{\mathbf{p}}_{size}$. We apply mask $\hat{\mathbf{p}}_{size}$ on features $\mathbf{f}^{q}$ to get $\mathbf{f}^{q}_{size}\in\mathbb{R}^{n\times\mathrm{D}}$. Then, we compute the cosine similarity between the selected query and support features and average it along the first dimension.
\begin{equation}
    \mathbf{p}^{rev}_{size}=\mathrm{mean}(\mathrm{cos}<\mathbf{f}^{q}_{size}, {\mathbf{f}^{s}}>, \mathrm{dim}=0).
\end{equation}
This operation gives us semantic priors $\mathbf{p}^{rev}_{size}$ on the support image. When $\mathbf{p}_{size}$ contains much noise, the reversely transferred target distribution on support image will highlight more irrelevant regions outside the polyp area, see the box in \cref{fig:sel}. Hence we set threshold for $\mathbf{p}^{rev}_{size}$ with $\tau$ and calculate the confidence Intersection over Union ($c\mathrm{IoU}$) between $\mathbf{p}^{rev}_{size}$ and $\mathbf{m}^{s}_{r}$. Different from conventional IoU, the intersection part of $c\mathrm{IoU}$ is multiplied with $\mathbf{p}^{rev}_{size}$ probability, which better measures the confidence of the target probability distribution. The ultimate adaptive weight is calculated as follows:
\begin{equation}
    \omega_{sz}=\frac{c\mathrm{IoU}_{sz}}{\sum_{sz}{c\mathrm{IoU}_{sz}}}, \: sz\in(ori, xl, xs).
\end{equation}
If $c\mathrm{IoU}_{sz}$ is low for a certain lesion size, the prior generated by the size is biased, and less credit $\omega_{sz}$ is given to the corresponding prior. The final prior is dynamically fused with a weighted combination of scale-cascaded priors:
\begin{equation}
    \mathbf{p}_{avg}=\sum_{sz}\omega_{sz}\mathbf{p}_{sz},\: sz\in(ori, xl, xs).
\end{equation}
\subsection{Euclidean Prompt Evolution}
\label{subsec:epe}
Given the final prior $\mathbf{p}_{avg}$, a self-prompting algorithm is needed for an end-to-end SAM-based segmentation framework. Previous methods either solely choose the highest response point \cite{zhang2023personalize} or entirely pull all points \cite{liu2023matcher} into the SAM decoder. The former inevitably focuses on a tiny part of the polyp area, while the latter contains much noise when the number of points exceeds limits. It is a well-recognized limitation that when input with excessive prompt points, SAM's performance deteriorates \cite{githubIssue95}. We propose the EPE module that fully utilizes the nature of SAM as an interactive segmentation model to achieve a superior balance between prompt quantity and segmentation accuracy.
\begin{algorithm}[t]
\caption{Euclidean Prompt Evolution Segmentation}
\label{alg:epe}
\begin{algorithmic}[1]

\STATE \textbf{Input:} query image $\mathbf{x}^{q}$, prior $\mathbf{p}$, tight prior thresh $\vartheta_{t}$, loose prior thresh $\vartheta_{l}$, score thresh $\theta$, negative area thresh $\eta$
\STATE \textbf{Output:} query mask $\mathbf{M}_{q}$

\STATE \textbf{Function} \textsc{EucSeg} ($\hat{\mathbf{p}}_{in}$: prior input, $\mathbf{L}$: prompt list)
\STATE \quad $\mathbf{\sigma}_{coor}=\mathrm{EDT}(\hat{\mathbf{p}}_{in})$ 
\STATE \quad $\mathbf{L}.$append($\mathbf{\sigma}_{coor}$)
\STATE \quad $\mathbf{M}, \:iou = \mathrm{SAM}(\mathbf{x}^{q}, \mathbf{L})$
\STATE \quad $cov = \mathrm{coverage}(\mathbf{M}, \hat{\mathbf{p}}_{t})$ \;
\STATE \quad \textbf{return} $\mathbf{M},\: cov,\: iou$
\vspace{1em} 
\STATE $\hat{\mathbf{p}}_{t}, \:\hat{\mathbf{p}}_{l} \leftarrow \mathbf{p} > \vartheta_{t}, \: \mathbf{p} > \vartheta_{l}$
\STATE $\mathbf{L}\leftarrow\emptyset$
\STATE $\mathbf{M}_{0},\: cov_{0},\: iou_{0} = $ \textsc{EucSeg}($\hat{\mathbf{p}}_{t}, \mathbf{L}$) \;
\STATE $done \leftarrow (cov_{0}\geq\theta)\:\textbf{and}\:(iou_{0}\geq\theta)$
\STATE $i \leftarrow 1$
\WHILE{\textbf{not} $done$}
    \IF{$cov_{i-1}<\theta$}
    \STATE $\hat{\mathbf{p}}_{i} = (\neg \mathbf{M}_{i-1})\cap \hat{\mathbf{p}}_{t}$;
    \ELSE
    \STATE $\hat{\mathbf{p}}_{i} = (\neg \mathbf{M}_{i-1})\cap \hat{\mathbf{p}}_{l}$;
    \ENDIF
    \STATE $\mathbf{M}_{i},\: cov_{i},\: iou_{i} = $ \textsc{EucSeg}($\hat{\mathbf{p}}_{i}, \mathbf{L}$) \;
    \STATE $i \leftarrow i + 1$
    \IF{$\mathrm{count}(\mathbf{M}_{i}\cap \neg \hat{\mathbf{p}}_{l})\geq\eta$}
    \STATE $\mathrm{del}\:\mathbf{M}_{i-1}$
    \STATE $\hat{\mathbf{p}}_{i} = \mathbf{M}_{i}\cap \neg \hat{\mathbf{p}}_{l}$;
    \STATE $\mathbf{M}_{i},\: cov_{i},\: iou_{i} = $ \textsc{EucSeg}($\hat{\mathbf{p}}_{i}, \mathbf{L}$) \;
    \STATE $i \leftarrow i + 1$
    \ENDIF
    \STATE $done\leftarrow(cov_{i}\geq\theta)\:\textbf{and}\:(iou_{i}\geq\theta)$
\ENDWHILE
\STATE \textbf{return} $\mathbf{M}_{q}=\bigcup_{j=0}^{i-1} \mathbf{M}_{j}$

\end{algorithmic}
\end{algorithm}

The core of EPE \cref{alg:epe} is \textsc{EucSeg} function to identify optimal prompt points. For an input prior $\hat{\mathbf{p}}_{in}$, it performs Euclidean Distance Transform (EDT) on the binary image and determines the point $\mathbf{\sigma}_{coor}$ with maximum distance from background pixels. Prompted with $\mathbf{\sigma}_{coor}$, SAM2 outputs target mask and mask iou prediction. \textsc{EucSeg} then calculates the coverage of the output mask over the prior and returns it with SAM outputs. The common practice to find the geometric prompt is calculating the bounding box center (BBC) of non-zero regions \cite{ayzenberg2024protosam}. However, we argue that the BBC can be skewed by irregular regions like thin and narrow areas. For example, in 3rd round prompting of \cref{fig:frame}, if BBC is employed, it will be located in the background due to the spike-like noise. In contrast, EDT center will focus more on primary object regions.

To initialize the EPE algorithm, we generate dual-threshold priors $\hat{\mathbf{p}}_{t},\hat{\mathbf{p}}_{l}$ using threshold $\vartheta_{t}, \vartheta_{l}$. The initial \textsc{EucSeg} is performed on highly-confident prior $\hat{\mathbf{p}}_{t}$. We evaluate the prediction by checking whether both $cov$ and $iou$ exceed the score threshold $\theta$. If coverage is low, it means SAM fails to encompass the confident regions. Hence we generate a prompt in the remaining areas of $\hat{\mathbf{p}}_{t}$ where the last-round mask does not cover. If coverage is high but $iou$ prediction is low, it indicates that $\hat{\mathbf{p}}_{t}$ does not provide a complete target prior. In that case, we expand to the loosely-threshed $\hat{\mathbf{p}}_{l}$ and run the \textsc{EucSeg} function on the remaining region of it. Note that we maintain a cumulative prompt list for records $\mathbf{L}$. Every time a new prompt is calculated, it is concatenated with previous prompts and feed-forward to the SAM predictor, and that is where the `evolution' in the algorithm name comes from. If the newly generated mask extends too much beyond $\hat{\mathbf{p}}_{l}$, we deem it as a noisy prediction and delete the prediction. For remediation, we add a negative prompt on noisy regions outside $\hat{\mathbf{p}}_{l}$. To avoid a dead loop, we jump out of the loop when more than five rounds of prompting have been performed. With the EPE algorithm, we ensure the comprehensive utilization of semantic information while maintaining segmentation accuracy through iterative validation.


\section{Experiments}
\subsection{Datasets and Evaluation Metrics}
\begin{table*}[t]
\centering
\caption{Quantitative performance comparison of the proposed method against state-of-the-art methods on Kvasir \cite{jha2020kvasir}, PolypGen \cite{ali2023multi}, CVC-ClinicDB \cite{bernal2015wm}, and CVC-ColonDB \cite{tajbakhsh2015automated} dataset. `Oracle' denotes randomly choosing three prompt points from the test ground truth labels.}
\begin{tabular}{l|c c|c c|c c|c c|c c|c c}
\hline
\multirow{3}{*}{Method} & \multicolumn{2}{c|}{Kvasir} & \multicolumn{6}{c|}{PolypGen} & \multicolumn{2}{c|}{CVC-ClinicDB} & \multicolumn{2}{c}{CVC-ColonDB}\\
\cline{2-13}
 & \multirow[b]{2}{*}{IoU} & \multirow[b]{2}{*}{Dice} & \multicolumn{2}{c|}{Center 1} & \multicolumn{2}{c|}{Center 2} & \multicolumn{2}{c|}{Center 3} & \multirow[b]{2}{*}{IoU} & \multirow[b]{2}{*}{Dice} & \multirow[b]{2}{*}{IoU} & \multirow[b]{2}{*}{Dice} \\
\cline{4-9}
 & & & IoU & Dice & IoU & Dice & IoU & Dice & & & & \\
\hline
\textcolor{gray}{\textit{Few-shot Spec}} &  &  &  &  &  &  &  &  &  &  &  &  \\
IFA\cite{nie2024cross} & 35.03 & 49.35 & 29.31 & 39.84 & 38.42 & 51.69 & 28.36 & 40.65 & 29.17 & 42.06 & 20.30 & 31.02 \\
\hline
\textcolor{gray}{\textit{In Context Seg}} &  &  &  &  &  &  &  &  &  &  &  &  \\
SegGPT\cite{wang2023seggpt} & 58.95 & 68.53 & 57.51 & 66.68 & 51.35 & 58.89 & 45.21 & 54.14 & 46.01 & 55.77 & 43.86 & 52.62 \\
\hline
\textcolor{gray}{\textit{One-shot SAM}} &  &  &  &  &  &  &  &  &  &  &  &  \\
\textcolor{gray}{Oracle\cite{ravi2024sam}}& \textcolor{gray}{71.88} & \textcolor{gray}{81.45} & \textcolor{gray}{69.21} & \textcolor{gray}{78.84} & \textcolor{gray}{72.09} & \textcolor{gray}{80.23} & \textcolor{gray}{78.22} & \textcolor{gray}{86.41} & \textcolor{gray}{74.30} & \textcolor{gray}{83.25} & \textcolor{gray}{74.84} & \textcolor{gray}{83.90} \\
EviPrompt\cite{eviprompt} & 18.64 & 27.51 & 8.09 & 13.11 & 18.93 & 25.49 & 7.70 & 13.05 & 19.84 & 26.83 & 8.06 & 13.23 \\
SEGIC\cite{meng2025segic} & 20.19 & 30.23 & 12.09 & 19.36 & 12.66 & 20.05 & 13.69 & 21.37 & 14.26 & 22.57 & 10.75 & 17.15 \\
PerSAM\cite{zhang2023personalize} & 57.97 & 67.76 & 46.29 & 53.06 & 58.47 & 64.46 & 13.81 & 17.99 & 51.27 & 58.55 & 50.30& 56.00 \\
PerSAM-f\cite{zhang2023personalize} & 42.95 & 51.40 & 43.47 & 50.09 & 46.19 & 52.61 & 8.99 & 11.97 & 40.97 & 48.31 & 33.90 & 38.24 \\
Matcher\cite{liu2023matcher} & \underline{65.49} & \underline{75.35} & \underline{61.76} & \underline{71.49} & \underline{65.15} & \underline{73.13} & \underline{65.03} & \underline{73.78} & \underline{61.94} & \underline{71.27} & \underline{54.92} & 63.06 \\
ProtoSAM\cite{ayzenberg2024protosam}& 64.27 & 75.13 & 59.15 & 68.82 & 55.28 & 64.44 & 56.42 & 66.40 & 59.85 & 69.77 & 53.79 & \underline{63.31} \\
\hline
OP-SAM(ours)& \textbf{76.93} & \textbf{84.53} & \textbf{68.68} & \textbf{77.21} & \textbf{68.65} & \textbf{75.51} & \textbf{71.44} & \textbf{77.88} & \textbf{67.75} & \textbf{75.56} & \textbf{60.81} & \textbf{68.06} \\
\hline
\end{tabular}
\label{tab:comparison}
\end{table*}

We validate the effectiveness of our method on five public polyp datasets. \textbf{Kvasir} \cite{jha2020kvasir}, \textbf{CVC-ClinicDB} \cite{bernal2015wm} and \textbf{CVC-ColonDB} \cite{tajbakhsh2015automated} are commonly used polyp segmentation datasets with variable-size polyp targets. To test the robustness and generalization capability, we adopt \textbf{PolypGen} \cite{ali2023multi} as the testbed. The dataset is collected from five different centers, and domain gaps exist between centers. To comprehensively evaluate our methods in practical medical scenarios, we perform experiments on self-curated extreme-size polyp segmentation referred to \cite{jain2023coinnet}, a more challenging but clinically meaningful task. Images with polyp coverage $\leq3\%$ or $\geq50\%$ of the entire image areas are collected from the Kvasir, and we call this set \textbf{Kvasir-H}. In addition, we conducted experiments on the \textbf{Piccolo} 
 \cite{sanchez2020piccolo} dataset, which contains more infiltrative, flat polyps that are difficult to detect with the human eyes. For fair comparison, we randomly select one image as the support and keep the support image fixed for all methods. We employ  Dice coefficient (Dice) and Intersection over Union (IoU) for evaluation.

\subsection{Implementation Details}
We use DINOv2 \cite{oquab2023dinov2} with a ViT-L/14 as the image feature encoder for prior generation and SAM2 \cite{ravi2024sam} with \textit{sam2-hiera-large} as the mask generator. We upgrade the mask generator of previous methods to SAM2 for fair comparison. The input size is set as $560\times560$ for DINOv2 and $1024\times1024$ for SAM2. For lesion scaling, we cut out the polyp with the mask, resize it and paste it back into the original image at the exact location. The blank gap between the zoomed-out polyp and original polyp is filled by inpainting. In the EPE module, the tight prior threshold $\vartheta_{t}$ and loose prior threshold $\vartheta_{l}$ are set as 0.7 and 0.5. The score thresh $\theta$ is set as 0.85 for confident prediction.

\subsection{Comparison with State-of-the-art Methods}
\noindent{\textbf{Generalized Polyp Segmentation.}} We compare our methods against specialist cross-domain few-shot segmentation models IFA \cite{nie2024cross}. As a special case for in-context learning, we compare our method against a generalized in-context segmentation framework SegGPT \cite{wang2023seggpt}. For SAM-based methods, our method is compared against state-of-the-art (SOTA) one-shot segmentation methods, including EviPrompt \cite{eviprompt}, SEGIC \cite{meng2025segic}, PerSAM \cite{zhang2023personalize}, ProtoSAM \cite{ayzenberg2024protosam}, Matcher \cite{liu2023matcher}. In addition, we perform an `oracle' experiment, where we randomly select three prompt points from the ground truth (GT) mask for each query image. The main results are shown in \cref{tab:comparison}. OP-SAM demonstrates superior performance over SOTA across all datasets, benefiting from its exhaustive semantic mining and recursive prompting algorithm. Specifically, on the Kvasir dataset, our method outperforms the second-best performer by 11.44\% in IoU and Dice. Surprisingly, our OP-SAM even surpasses the `oracle' on the Kvasir dataset. The superiority can be attributed to our EPE prompting, which ensures prior coverage through the `evaluate-prompt-segment' loop. When randomly sampling prompts from GT, the chosen points may be clustered in a tiny region. The outcome once again emphasizes the importance of designing a sophisticated prompting method. Otherwise, segmentation results could be unsatisfactory even though the GT is given.

\begin{figure*}[t]
  \centering
   \includegraphics[width=0.9\textwidth]{{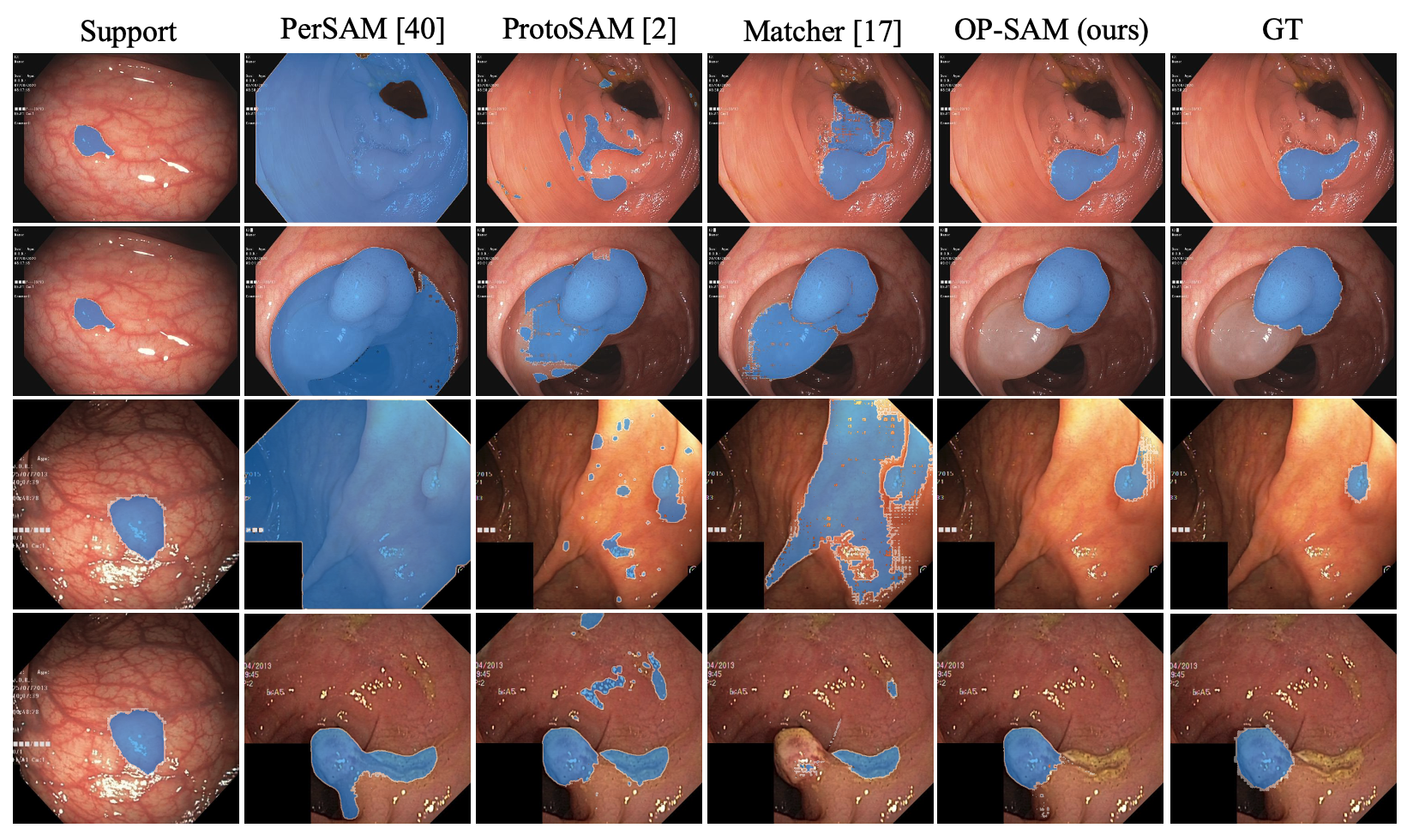}}
    \caption{Qualitative comparison of OP-SAM against state-of-the-art methods on PolypGen \cite{ali2023multi} and Kvasir \cite{jha2020kvasir} datasets.}
   \label{fig:comp}
\end{figure*}

\begin{figure*}[t]
  \centering
   \includegraphics[width=0.9\textwidth]{{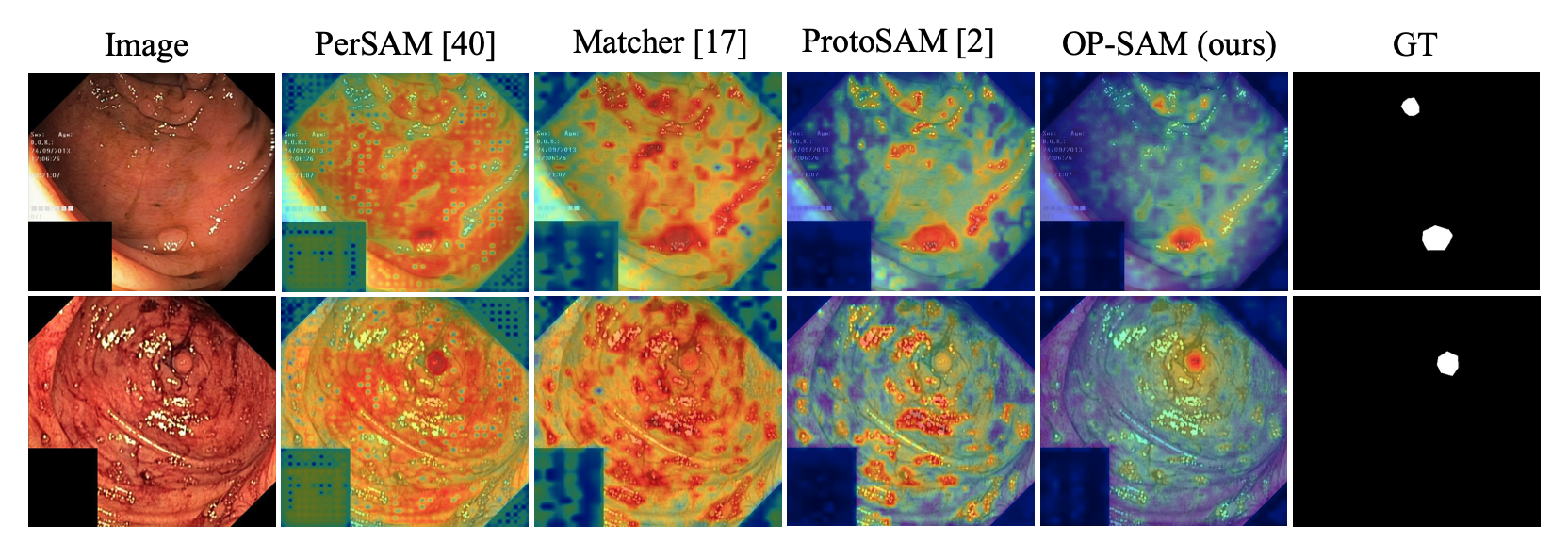}}
    \caption{Visualization of prior of our method and state-of-the-art methods.}
   \label{fig:prior_comp}
\end{figure*}

\noindent{\textbf{Challenging Cases Experiments.}} To  demonstrate the robustness of our method, we conduct comparison on two challenging datasets Kvasir-H and Piccolo. The former contains manually selected samples with extreme sizes, and the latter encompasses more plaque-like hard-to-find polyps. The results are shown in \cref{tab:spolyps}. Our approach outperforms the second-best method on Kvasir-H by 10\% in IoU, which demonstrates the effectiveness of our scale-cascade prior fusion. Our approach closely approximates the results of the oracle, demonstrating the precision of prior.

\begin{table}[t]
\centering
\caption{Performance comparison of OP-SAM against state-of-the-art methods on challenging datasets.}
\begin{tabular}{l|c c|c c}
\hline
\multirow{2}{*}{Method} & \multicolumn{2}{c|}{Kvasir-H} & \multicolumn{2}{c}{Piccolo}\\
\cline{2-5}
 & IoU & Dice & IoU & Dice \\
\hline
IFA\cite{nie2024cross} & 22.29 & 33.32 & 25.59 & 36.24 \\
SegGPT\cite{wang2023seggpt} & 29.01 & 37.87 & 41.84 & 49.79 \\
\hline
\textcolor{gray}{Oracle\cite{ravi2024sam}}& \textcolor{gray}{61.03} & \textcolor{gray}{71.85} & \textcolor{gray}{72.82} & \textcolor{gray}{80.29} \\
PerSAM\cite{zhang2023personalize} & \underline{47.05} & \underline{53.59} & 51.17 & 57.54 \\
PerSAM-f\cite{zhang2023personalize} & 25.45 & 30.64 & 47.84 & 53.08 \\
Matcher\cite{liu2023matcher} & 33.23 & 45.00 & \underline{59.43} & \underline{68.19} \\
ProtoSAM\cite{ayzenberg2024protosam}& 36.06 & 48.94 & 47.69 & 58.39 \\
\hline
OP-SAM(ours)& \textbf{57.31} & \textbf{67.56} & \textbf{65.48} & \textbf{72.26} \\
\hline
\end{tabular}

\label{tab:spolyps}
\end{table}

\begin{table}[t]
\centering
\caption{Ablation studies of key modules on Kvasir Dataset. Random 3 points are sampled from the prior if EPE prompting is not applied.}
\begin{tabular}{c | c c c | c c }
\hline
 \multirow{2}{*}{Method} & \multicolumn{3}{c|}{Modules} & \multirow{2}{*}{IoU} & \multirow{2}{*}{Dice} \\
\cline{2-4}
 & CPG & SPF & EPE &  & \\
\hline
\multirow{3}{*}{OP-SAM} & \ding{52} & & & 67.12 & 76.43\\
& \ding{52} & \ding{52}& & 69.05 & 78.07 \\
& \ding{52} & \ding{52}& \ding{52} & \textbf{76.93} & \textbf{84.53}  \\
\hline
\end{tabular}

\label{tab:key_aba}
\end{table}

\noindent{\textbf{Qualitative Results}} \cref{fig:comp} visualizes some challenging cases for qualitative comparison.  In cases where polyps infiltrate the colon wall, our method achieves precise segmentation while competing approaches fail to identify target lesions (row 1). The method exhibits particular robustness when handling significant size disparities between support and query images (row 2-3). Furthermore, OP-SAM demonstrates superior discriminative ability in the presence of intestinal secretions and debris (row 4). Prior map visualization \cref{fig:prior_comp} reveals more precise target localization with notably reduced noise compared to existing approaches.

\subsection{Ablation Studies}
Ablation studies \cref{tab:key_aba} evaluate the combined contributions of the key components. Results demonstrate the critical importance of both accurate prior generation and progressive prompting strategies.  The integration of these components produces a synergistic effect, yielding optimal performance across both IoU and Dice metrics. 

\noindent{\textbf{CPG Analysis.}} The effectiveness of different prior generation components \cref{tab:prior_aba} is evaluated by thresholding the prior and calculating IoU and Dice metrics against ground truth. The base method is cross-correlation generation. When equipped with self-correlation, it yields a substantial 30\% accuracy improvement.

\begin{table}[t]
\centering
\caption{Ablation studies of prior generation methods on Kvasir dataset, CG denotes Cross-correlation Generation, SR denotes Self-correlation Refinement.}

\begin{tabular}{c | c c | c c c}
\hline

 \multirow{2}{*}{Method} & \multicolumn{2}{c|}{CPG} & \multicolumn{3}{c}{Prior} \\
\cline{2-6}
 & CG & SR & IoU & Dice & roc-auc\\
\hline
PerSAM\cite{zhang2023personalize} & - & - & 0.23 & 0.47 & -\\
ProtoSAM\cite{ayzenberg2024protosam} & -  & - & 55.12 & 68.81 & 93.89\\
\hline
\multirow{2}{*}{OP-SAM} & \ding{52} & & 31.88 & 43.82 & 95.81\\
& \ding{52} & \ding{52}& \textbf{58.30} & \textbf{70.69} & \textbf{98.37}\\
\hline
\end{tabular}

\label{tab:prior_aba}
\end{table}

\begin{table}[t]
\centering
\caption{Ablation studies of support types on Kvasir dataset. `aug' denotes lesion scaling augmentation, `Avg' denotes mean average and SPF is our adaptive fusion.}
\begin{tabular}{c |c |c c}
\hline
Method & Support Type & IoU  & Dice\\
\hline
\multirow{5}{*}{OP-SAM} & 3 images + Avg & 37.88 & 49.61 \\ 
 & 3 images + SPF & 57.23 & 70.34 \\ 
 & 5 images + Avg & 44.53 & 57.17 \\ 
 & 1 image + 2 augs + Avg & 48.67 & 61.09 \\
 & 1 image + 2 augs + SPF & \textbf{61.47} & \textbf{73.71} \\ 
\hline
\end{tabular}

\label{tab:supp_num_aba}
\end{table}

\noindent{\textbf{Support Configuration Study.}} Investigation of support image quantity \cref{tab:supp_num_aba} reveals interesting patterns in prior quality. The evaluation is performed in the same way of \cref{tab:prior_aba}. Our target-scaling approach with a single mask achieves 4\% increase in IoU compared to 3-shot methods using SPF. The scale cascaded multiple prior is ineffective when independently used. This is because different support sizes generate prior highlighting different regions, and roughly averaging them is bound to introduce noise. However, with SPF it significantly enhances prior accuracy. Notably, increasing support image count without a proper fusion strategy does not yield significant improvements. This finding emphasizes the importance of fusion methodology over support quantity.

\noindent{\textbf{Prompting Strategy Evaluation.}} To test the superiority of our EPE module, we maintain our prior generation fixed and ablate the prompting method. The results is shown in \cref{tab:prompt_aba}. We select the highest and lowest probability points, following \cite{zhang2023personalize}. Referring to \cite{ayzenberg2024protosam}, we we input the highest probability point, the center of the non-zero region bounding box and the bounding box itself into SAM2. Our EPE algorithm outperforms all methods mentioned above. We further change the prompt point in EPE from Euclidean Distance Transformed Center to Bounding Box Center. The prompt evolution collapses as we analyzed in Sec. \ref{subsec:epe}. This ablation highlights the novelty of EDT in simulating interactive feedback for SAM2.
\begin{figure}[t]
  \centering
   \includegraphics[width=0.45\textwidth]{{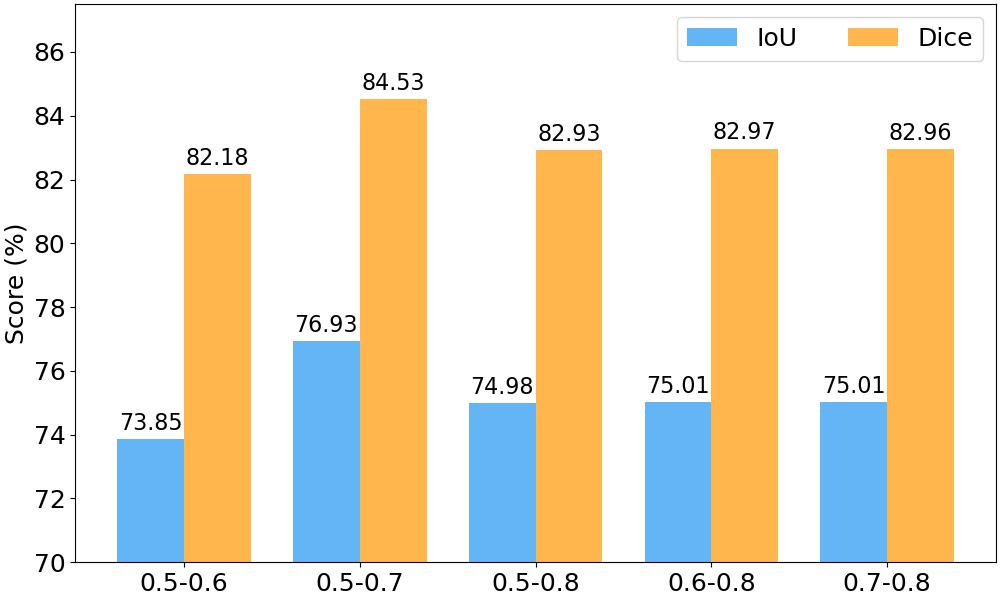}}
    \caption{Ablation studies of tight and loose threshold. The x label is `$\vartheta_l$-$\vartheta_t$' (loose and tight threshold). }
   \label{fig:thresh_abla}
\end{figure}

\begin{table}[t]
\centering
\caption{Ablation studies of prompting methods on Kvasir dataset}
\begin{tabular}{c |c |c c}
\hline
Method & Prompting Method & IoU  & Dice\\
\hline
\multirow{4}{*}{OP-SAM} & top first and last & 60.49 & 69.13 \\
 & 1 top + 1 center + bbox& 70.54 & 79.51 \\
 & BB center evolution & 58.04 & 64.85 \\
 & EPE center evolution & \textbf{76.93} & \textbf{84.53} \\
\hline
\end{tabular}

\label{tab:prompt_aba}
\end{table}

\begin{table}[t]
\centering
\caption{Ablation studies of different support images}
\begin{tabular}{ c | c | c | c | c }
\hline
 & PerSAM & Matcher & ProtoSAM & Ours \\
\hline
$\mathrm{avg}_{iou}\uparrow$ & 44.00 & 66.35 & 66.19 & \textbf{75.81} \\
$\mathrm{var}_{iou}\downarrow$ & 12.57 & 2.55 & 2.80 & \textbf{0.48}\\
\hline
\end{tabular}

\label{tab:supp_aba}
\end{table}

\noindent{\textbf{Support Image Ablations}} We conducted an additional ablation study by randomly selecting five different support images and computing the mean and variance of IoU from different methods. As shown in \cref{tab:supp_aba}, our method not only maintains nearly a 10\% performance gain over the SOTA but also achieves the lowest deviation across different support images. This further demonstrates that our method is robust to the selection of the support image.

\noindent{\textbf{Hyper-parameter Analysis.}} Our investigation of dual threshold parameters in EPE ($\vartheta_{t}$ - tight threshold, $\vartheta_{l}$ - loose threshold) reveals optimal performance at $\vartheta_{t}=0.7$ and $\vartheta_{l}=0.5$, achieving 76.93\% IoU. Both large gaps (0.5-0.8) and narrow ranges (0.7-0.8) lead to reduced performance. The results demonstrate that moderately spaced thresholds provide the best balance between precision and robustness.

\section{Conclusion}
We present OP-SAM, a novel training-free framework for polyp segmentation that operates with minimal supervision, requiring only a single image-label pair. Our approach advances the state-of-the-art through three key innovations: CPG utilizing cross and within-image relationships to accurately transfer the labels from support image to query image, SPF adaptively fusing scale-aware priors and excluding noisy prior with reverse transfer, and iterative prompting mechanism EPE based on last-round feedback. Extensive evaluation across multiple datasets demonstrates that OP-SAM achieves superior accuracy and robustness compared to existing methods. The robust performance across multiple datasets suggests that our method provides a promising approach for generalized polyp segmentation without excessive additional costs.

\vspace{2ex}
\noindent{\textbf{Acknowledgement.}} This work is supported by Shenzhen Key Laboratory of Robotics Perception and Intelligence (ZDSYS20200810171800001), and the High level of special funds (G03034K003) from Southern University of Science and Technology, Shenzhen, China.

{
    \small
    \bibliographystyle{ieeenat_fullname}
    \bibliography{main}

\begin{thebibliography}{41}
\providecommand{\natexlab}[1]{#1}
\providecommand{\url}[1]{\texttt{#1}}
\expandafter\ifx\csname urlstyle\endcsname\relax
  \providecommand{\doi}[1]{doi: #1}\else
  \providecommand{\doi}{doi: \begingroup \urlstyle{rm}\Url}\fi

\bibitem[Ali et~al.(2023)Ali, Jha, Ghatwary, Realdon, Cannizzaro, Salem, Lamarque, Daul, Riegler, Anonsen, et~al.]{ali2023multi}
Sharib Ali, Debesh Jha, Noha Ghatwary, Stefano Realdon, Renato Cannizzaro, Osama~E Salem, Dominique Lamarque, Christian Daul, Michael~A Riegler, Kim~V Anonsen, et~al.
\newblock A multi-centre polyp detection and segmentation dataset for generalisability assessment.
\newblock \emph{Scientific Data}, 10\penalty0 (1):\penalty0 75, 2023.

\bibitem[Ayzenberg et~al.(2024)Ayzenberg, Giryes, and Greenspan]{ayzenberg2024protosam}
Lev Ayzenberg, Raja Giryes, and Hayit Greenspan.
\newblock Protosam-one shot medical image segmentation with foundational models.
\newblock \emph{arXiv preprint arXiv:2407.07042}, 2024.

\bibitem[Bernal et~al.(2015)Bernal, S{\'a}nchez, Fern{\'a}ndez-Esparrach, Gil, Rodr{\'\i}guez, and Vilari{\~n}o]{bernal2015wm}
Jorge Bernal, F~Javier S{\'a}nchez, Gloria Fern{\'a}ndez-Esparrach, Debora Gil, Cristina Rodr{\'\i}guez, and Fernando Vilari{\~n}o.
\newblock Wm-dova maps for accurate polyp highlighting in colonoscopy: Validation vs. saliency maps from physicians.
\newblock \emph{Computerized medical imaging and graphics}, 43:\penalty0 99--111, 2015.

\bibitem[Chen et~al.(2021)Chen, Xie, Yao, Wang, Shen, Tang, and Zhang]{chen2021semantically}
Tao Chen, Guo-Sen Xie, Yazhou Yao, Qiong Wang, Fumin Shen, Zhenmin Tang, and Jian Zhang.
\newblock Semantically meaningful class prototype learning for one-shot image segmentation.
\newblock \emph{IEEE Transactions on Multimedia}, 24:\penalty0 968--980, 2021.

\bibitem[Gao et~al.(2022)Gao, Xiao, Yin, Liu, and Shi]{gao2022mutually}
Honghao Gao, Junsheng Xiao, Yuyu Yin, Tong Liu, and Jiangang Shi.
\newblock A mutually supervised graph attention network for few-shot segmentation: The perspective of fully utilizing limited samples.
\newblock \emph{IEEE Transactions on neural networks and learning systems}, 35\penalty0 (4):\penalty0 4826--4838, 2022.

\bibitem[Jain et~al.(2023)Jain, Atale, Gupta, Mishra, Seal, Ojha, Jaworek-Korjakowska, and Krejcar]{jain2023coinnet}
Samir Jain, Rohan Atale, Anubhav Gupta, Utkarsh Mishra, Ayan Seal, Aparajita Ojha, Joanna Jaworek-Korjakowska, and Ondrej Krejcar.
\newblock Coinnet: A convolution-involution network with a novel statistical attention for automatic polyp segmentation.
\newblock \emph{IEEE Transactions on Medical Imaging}, 42\penalty0 (12):\penalty0 3987--4000, 2023.

\bibitem[Jha et~al.(2020)Jha, Smedsrud, Riegler, Halvorsen, De~Lange, Johansen, and Johansen]{jha2020kvasir}
Debesh Jha, Pia~H Smedsrud, Michael~A Riegler, P{\aa}l Halvorsen, Thomas De~Lange, Dag Johansen, and H{\aa}vard~D Johansen.
\newblock Kvasir-seg: A segmented polyp dataset.
\newblock In \emph{MultiMedia modeling: 26th international conference, MMM 2020, Daejeon, South Korea, January 5--8, 2020, proceedings, part II 26}, pages 451--462. Springer, 2020.

\bibitem[Jha et~al.(2024)Jha, Tomar, Sharma, and Bagci]{jha2024transnetr}
Debesh Jha, Nikhil~Kumar Tomar, Vanshali Sharma, and Ulas Bagci.
\newblock Transnetr: transformer-based residual network for polyp segmentation with multi-center out-of-distribution testing.
\newblock In \emph{Medical Imaging with Deep Learning}, pages 1372--1384. PMLR, 2024.

\bibitem[Lang et~al.(2022)Lang, Cheng, Tu, and Han]{lang2022learning}
Chunbo Lang, Gong Cheng, Binfei Tu, and Junwei Han.
\newblock Learning what not to segment: A new perspective on few-shot segmentation.
\newblock In \emph{Proceedings of the IEEE/CVF conference on computer vision and pattern recognition}, pages 8057--8067, 2022.

\bibitem[Lei et~al.(2022)Lei, Zhang, He, Chen, Du, and Lu]{lei2022cross}
Shuo Lei, Xuchao Zhang, Jianfeng He, Fanglan Chen, Bowen Du, and Chang-Tien Lu.
\newblock Cross-domain few-shot semantic segmentation.
\newblock In \emph{European Conference on Computer Vision}, pages 73--90. Springer, 2022.

\bibitem[Li et~al.(2023)Li, Zhang, Sun, Zou, Liu, Yang, Li, Zhang, and Gao]{li2023semantic}
Feng Li, Hao Zhang, Peize Sun, Xueyan Zou, Shilong Liu, Jianwei Yang, Chunyuan Li, Lei Zhang, and Jianfeng Gao.
\newblock Semantic-sam: Segment and recognize anything at any granularity.
\newblock \emph{arXiv preprint arXiv:2307.04767}, 2023.

\bibitem[Li et~al.(2021)Li, Jampani, Sevilla-Lara, Sun, Kim, and Kim]{li2021adaptive}
Gen Li, Varun Jampani, Laura Sevilla-Lara, Deqing Sun, Jonghyun Kim, and Joongkyu Kim.
\newblock Adaptive prototype learning and allocation for few-shot segmentation.
\newblock In \emph{Proceedings of the IEEE/CVF conference on computer vision and pattern recognition}, pages 8334--8343, 2021.

\bibitem[Li et~al.(2022)Li, Xu, Zhang, Feng, Zhao, Zhang, Lu, and Gao]{li2022tccnet}
Xiaotong Li, Jilan Xu, Yuejie Zhang, Rui Feng, Rui-Wei Zhao, Tao Zhang, Xuequan Lu, and Shang Gao.
\newblock Tccnet: Temporally consistent context-free network for semi-supervised video polyp segmentation.
\newblock In \emph{IJCAI}, pages 1109--1115, 2022.

\bibitem[Li et~al.(2024)Li, Hu, and Yang]{li2024polyp}
Yuheng Li, Mingzhe Hu, and Xiaofeng Yang.
\newblock Polyp-sam: Transfer sam for polyp segmentation.
\newblock In \emph{Medical Imaging 2024: Computer-Aided Diagnosis}, pages 759--765. SPIE, 2024.

\bibitem[Lin et~al.(2023)Lin, Chen, Wang, Wu, Li, Lin, Liu, and He]{lin2023clip}
Yuqi Lin, Minghao Chen, Wenxiao Wang, Boxi Wu, Ke Li, Binbin Lin, Haifeng Liu, and Xiaofei He.
\newblock Clip is also an efficient segmenter: A text-driven approach for weakly supervised semantic segmentation.
\newblock In \emph{Proceedings of the IEEE/CVF Conference on Computer Vision and Pattern Recognition}, pages 15305--15314, 2023.

\bibitem[Liu et~al.(2022)Liu, Liu, Yao, and Han]{liu2022intermediate}
Yuanwei Liu, Nian Liu, Xiwen Yao, and Junwei Han.
\newblock Intermediate prototype mining transformer for few-shot semantic segmentation.
\newblock \emph{Advances in Neural Information Processing Systems}, 35:\penalty0 38020--38031, 2022.

\bibitem[Liu et~al.(2023)Liu, Zhu, Li, Chen, Wang, and Shen]{liu2023matcher}
Yang Liu, Muzhi Zhu, Hengtao Li, Hao Chen, Xinlong Wang, and Chunhua Shen.
\newblock Matcher: Segment anything with one shot using all-purpose feature matching.
\newblock \emph{arXiv preprint arXiv:2305.13310}, 2023.

\bibitem[Lujiazho et~al.(2023)]{githubIssue95}
Lujiazho et~al.
\newblock Bad results with multiple point prompts.
\newblock \url{https://github.com/facebookresearch/segment-anything/issues/95}, 2023.
\newblock Accessed: 2024-11-09.

\bibitem[Ma et~al.(2024)Ma, He, Li, Han, You, and Wang]{ma2024segment}
Jun Ma, Yuting He, Feifei Li, Lin Han, Chenyu You, and Bo Wang.
\newblock Segment anything in medical images.
\newblock \emph{Nature Communications}, 15\penalty0 (1):\penalty0 654, 2024.

\bibitem[Meng et~al.(2025)Meng, Lan, Li, Alvarez, Wu, and Jiang]{meng2025segic}
Lingchen Meng, Shiyi Lan, Hengduo Li, Jose~M Alvarez, Zuxuan Wu, and Yu-Gang Jiang.
\newblock Segic: Unleashing the emergent correspondence for in-context segmentation.
\newblock In \emph{European Conference on Computer Vision}, pages 203--220. Springer, 2025.

\bibitem[Nie et~al.(2024)Nie, Xing, Zhang, Yan, Xiao, Tan, Kot, and Lu]{nie2024cross}
Jiahao Nie, Yun Xing, Gongjie Zhang, Pei Yan, Aoran Xiao, Yap-Peng Tan, Alex~C Kot, and Shijian Lu.
\newblock Cross-domain few-shot segmentation via iterative support-query correspondence mining.
\newblock In \emph{Proceedings of the IEEE/CVF Conference on Computer Vision and Pattern Recognition}, pages 3380--3390, 2024.

\bibitem[Oquab et~al.(2023)Oquab, Darcet, Moutakanni, Vo, Szafraniec, Khalidov, Fernandez, Haziza, Massa, El-Nouby, et~al.]{oquab2023dinov2}
Maxime Oquab, Timoth{\'e}e Darcet, Th{\'e}o Moutakanni, Huy Vo, Marc Szafraniec, Vasil Khalidov, Pierre Fernandez, Daniel Haziza, Francisco Massa, Alaaeldin El-Nouby, et~al.
\newblock Dinov2: Learning robust visual features without supervision.
\newblock \emph{arXiv preprint arXiv:2304.07193}, 2023.

\bibitem[Patel et~al.(2022)Patel, Li, and Wang]{patel2022fuzzynet}
Krushi~Bharatbhai Patel, Fengjun Li, and Guanghui Wang.
\newblock Fuzzynet: A fuzzy attention module for polyp segmentation.
\newblock In \emph{NeurIPS'22 Workshop on All Things Attention: Bridging Different Perspectives on Attention}, 2022.

\bibitem[Radford et~al.(2021)Radford, Kim, Hallacy, Ramesh, Goh, Agarwal, Sastry, Askell, Mishkin, Clark, et~al.]{radford2021clip}
Alec Radford, Jong~Wook Kim, Chris Hallacy, Aditya Ramesh, Gabriel Goh, Sandhini Agarwal, Girish Sastry, Amanda Askell, Pamela Mishkin, Jack Clark, et~al.
\newblock Learning transferable visual models from natural language supervision.
\newblock In \emph{International conference on machine learning}, pages 8748--8763. PMLR, 2021.

\bibitem[Rahman and Marculescu(2023)]{rahman2023medical}
Md~Mostafijur Rahman and Radu Marculescu.
\newblock Medical image segmentation via cascaded attention decoding.
\newblock In \emph{Proceedings of the IEEE/CVF Winter Conference on Applications of Computer Vision}, pages 6222--6231, 2023.

\bibitem[Rahman et~al.(2024)Rahman, Munir, Jha, Bagci, and Marculescu]{rahman2024pp}
Md~Mostafijur Rahman, Mustafa Munir, Debesh Jha, Ulas Bagci, and Radu Marculescu.
\newblock Pp-sam: Perturbed prompts for robust adaption of segment anything model for polyp segmentation.
\newblock In \emph{Proceedings of the IEEE/CVF Conference on Computer Vision and Pattern Recognition}, pages 4989--4995, 2024.

\bibitem[Ravi et~al.(2024)Ravi, Gabeur, Hu, Hu, Ryali, Ma, Khedr, R{\"a}dle, Rolland, Gustafson, et~al.]{ravi2024sam}
Nikhila Ravi, Valentin Gabeur, Yuan-Ting Hu, Ronghang Hu, Chaitanya Ryali, Tengyu Ma, Haitham Khedr, Roman R{\"a}dle, Chloe Rolland, Laura Gustafson, et~al.
\newblock Sam 2: Segment anything in images and videos.
\newblock \emph{arXiv preprint arXiv:2408.00714}, 2024.

\bibitem[S{\'a}nchez-Peralta et~al.(2020)S{\'a}nchez-Peralta, Pagador, Pic{\'o}n, Calder{\'o}n, Polo, Andraka, Bilbao, Glover, Saratxaga, and S{\'a}nchez-Margallo]{sanchez2020piccolo}
Luisa~F S{\'a}nchez-Peralta, J~Blas Pagador, Artzai Pic{\'o}n, {\'A}ngel~Jos{\'e} Calder{\'o}n, Francisco Polo, Nagore Andraka, Roberto Bilbao, Ben Glover, Cristina~L Saratxaga, and Francisco~M S{\'a}nchez-Margallo.
\newblock Piccolo white-light and narrow-band imaging colonoscopic dataset: A performance comparative of models and datasets.
\newblock \emph{Applied Sciences}, 10\penalty0 (23):\penalty0 8501, 2020.

\bibitem[Shi et~al.(2022)Shi, Wei, Zhang, Lu, Ning, Chen, Ma, and Zheng]{shi2022dense}
Xinyu Shi, Dong Wei, Yu Zhang, Donghuan Lu, Munan Ning, Jiashun Chen, Kai Ma, and Yefeng Zheng.
\newblock Dense cross-query-and-support attention weighted mask aggregation for few-shot segmentation.
\newblock In \emph{European Conference on Computer Vision}, pages 151--168. Springer, 2022.

\bibitem[Srivastava et~al.(2022)Srivastava, Jha, Chanda, Pal, Johansen, Johansen, Riegler, Ali, and Halvorsen]{msrf}
Abhishek Srivastava, Debesh Jha, Sukalpa Chanda, Umapada Pal, Håvard~D. Johansen, Dag Johansen, Michael~A. Riegler, Sharib Ali, and Pål Halvorsen.
\newblock Msrf-net: A multi-scale residual fusion network for biomedical image segmentation.
\newblock \emph{IEEE Journal of Biomedical and Health Informatics}, 26\penalty0 (5):\penalty0 2252--2263, 2022.

\bibitem[Tajbakhsh et~al.(2015)Tajbakhsh, Gurudu, and Liang]{tajbakhsh2015automated}
Nima Tajbakhsh, Suryakanth~R Gurudu, and Jianming Liang.
\newblock Automated polyp detection in colonoscopy videos using shape and context information.
\newblock \emph{IEEE transactions on medical imaging}, 35\penalty0 (2):\penalty0 630--644, 2015.

\bibitem[Tian et~al.(2023)Tian, Liu, Pang, Chen, Liu, Verjans, Singh, and Carneiro]{tian2023self}
Yu Tian, Fengbei Liu, Guansong Pang, Yuanhong Chen, Yuyuan Liu, Johan~W Verjans, Rajvinder Singh, and Gustavo Carneiro.
\newblock Self-supervised pseudo multi-class pre-training for unsupervised anomaly detection and segmentation in medical images.
\newblock \emph{Medical image analysis}, 90:\penalty0 102930, 2023.

\bibitem[Wang et~al.(2023{\natexlab{a}})Wang, Xu, Zhang, Islam, and Ren]{wang2023s}
An Wang, Mengya Xu, Yang Zhang, Mobarakol Islam, and Hongliang Ren.
\newblock S 2 me: Spatial-spectral mutual teaching and ensemble learning for scribble-supervised polyp segmentation.
\newblock In \emph{International Conference on Medical Image Computing and Computer-Assisted Intervention}, pages 35--45. Springer, 2023{\natexlab{a}}.

\bibitem[Wang et~al.(2023{\natexlab{b}})Wang, Chen, Ma, Wang, Fei, Shuai, Tang, Zhou, and Qin]{wang2023xbound}
Jiacheng Wang, Fei Chen, Yuxi Ma, Liansheng Wang, Zhaodong Fei, Jianwei Shuai, Xiangdong Tang, Qichao Zhou, and Jing Qin.
\newblock Xbound-former: Toward cross-scale boundary modeling in transformers.
\newblock \emph{IEEE Transactions on Medical Imaging}, 42\penalty0 (6):\penalty0 1735--1745, 2023{\natexlab{b}}.

\bibitem[Wang et~al.(2023{\natexlab{c}})Wang, Zhang, Cao, Wang, Shen, and Huang]{wang2023seggpt}
Xinlong Wang, Xiaosong Zhang, Yue Cao, Wen Wang, Chunhua Shen, and Tiejun Huang.
\newblock Seggpt: Towards segmenting everything in context.
\newblock In \emph{Proceedings of the IEEE/CVF International Conference on Computer Vision}, pages 1130--1140, 2023{\natexlab{c}}.

\bibitem[Wei et~al.(2021)Wei, Hu, Zhang, Li, Zhou, and Cui]{wei2021shallow}
Jun Wei, Yiwen Hu, Ruimao Zhang, Zhen Li, S~Kevin Zhou, and Shuguang Cui.
\newblock Shallow attention network for polyp segmentation.
\newblock In \emph{Medical Image Computing and Computer Assisted Intervention--MICCAI 2021: 24th International Conference, Strasbourg, France, September 27--October 1, 2021, Proceedings, Part I 24}, pages 699--708. Springer, 2021.

\bibitem[Xu et~al.(2024)Xu, Tang, Men, and Chen]{eviprompt}
Yinsong Xu, Jiaqi Tang, Aidong Men, and Qingchao Chen.
\newblock Eviprompt: A training-free evidential prompt generation method for adapting segment anything model in medical images.
\newblock \emph{IEEE Transactions on Image Processing}, 33:\penalty0 6204--6215, 2024.

\bibitem[Yang et~al.(2023)Yang, Chen, Feng, and Huang]{yang2023mianet}
Yong Yang, Qiong Chen, Yuan Feng, and Tianlin Huang.
\newblock Mianet: Aggregating unbiased instance and general information for few-shot semantic segmentation.
\newblock In \emph{Proceedings of the IEEE/CVF Conference on Computer Vision and Pattern Recognition}, pages 7131--7140, 2023.

\bibitem[Zhang et~al.(2021)Zhang, Kang, Yang, and Wei]{zhang2021few}
Gengwei Zhang, Guoliang Kang, Yi Yang, and Yunchao Wei.
\newblock Few-shot segmentation via cycle-consistent transformer.
\newblock \emph{Advances in Neural Information Processing Systems}, 34:\penalty0 21984--21996, 2021.

\bibitem[Zhang et~al.(2023)Zhang, Jiang, Guo, Yan, Pan, Ma, Dong, Gao, and Li]{zhang2023personalize}
Renrui Zhang, Zhengkai Jiang, Ziyu Guo, Shilin Yan, Junting Pan, Xianzheng Ma, Hao Dong, Peng Gao, and Hongsheng Li.
\newblock Personalize segment anything model with one shot.
\newblock \emph{arXiv preprint arXiv:2305.03048}, 2023.

\bibitem[Zhou et~al.(2023)Zhou, Zhou, He, Gong, Yang, Fu, and Shen]{zhou2023cross}
Tao Zhou, Yi Zhou, Kelei He, Chen Gong, Jian Yang, Huazhu Fu, and Dinggang Shen.
\newblock Cross-level feature aggregation network for polyp segmentation.
\newblock \emph{Pattern Recognition}, 140:\penalty0 109555, 2023.

\end{thebibliography}
}


\end{document}